\documentclass[journal]{IEEEtran}
\usepackage{cite}
\usepackage{hyperref}

\usepackage[utf8x]{inputenc}

\ifCLASSINFOpdf
  \usepackage[pdftex]{graphicx}
\else
\fi
\begin{document}
%
\title{COVID-19 growth prediction using multivariate long short term memory}
%
%
%

\author{Novanto Yudistira
        
\thanks{N. Yudistira is with Intelligent System Lab, Faculty of Computer Science, Brawijaya University, Indonesia
 e-mail: yudistira@ub.ac.id}
}
%
%

\markboth{Journal of \LaTeX\ Class Files,~Vol.~14, No.~8, August~2015}%
{Shell \MakeLowercase{\textit{et al.}}: Bare Demo of IEEEtran.cls for IEEE Journals}
%



\maketitle

\begin{abstract}

Coronavirus disease (COVID-19) spread forecasting is
an important task to track the growth of the pandemic. Existing
predictions are merely based on qualitative analyses and mathematical modeling. The use of available big
data with machine learning is still limited in COVID-19 growth
prediction even though the availability of data is abundance.
To make use of big data in the prediction using deep learning,
we use long short-term memory (LSTM) method to learn
the correlation of COVID-19 growth over time. The structure of an 
LSTM layer is searched heuristically until the best
validation score is achieved. First, we trained training data containing
confirmed cases from around the globe. We achieved favorable
performance compared with that of the recurrent neural network (RNN) method with a comparable low
validation error. The evaluation is conducted based on graph visualization and root mean squared error (RMSE). We found that it is not easy to achieve
the same quantity of confirmed cases over time. However,
LSTM provide a similar pattern between the actual cases and
prediction. In the future, our proposed prediction can be used for
anticipating forthcoming pandemics. The code is provided
here: \url{https://github.com/cbasemaster/lstmcorona}

\end{abstract}


%
\IEEEpeerreviewmaketitle

\section{Introduction}

\IEEEPARstart{T}{he COVID-19} outbreak first occurred in China and then gradually spread around the world. The factors that cause the outbreak are still in the discussion phase. However, many countries have been anticipating the transmission using social distancing and activity restrictions except Sweden \cite{paterlini}. Since then, not many predictions are available except qualitative and statistical analyses \cite{peng}\cite{roda}\cite{sajadi}\cite{bevenuto}. Although long short-term memory (LSTM) has been applied in various and diverse time-series topics, such as stock prediction, weather, and consumer, findings on the exact manifestations of COVID-19 are still limited. LSTM was used to predict the end of the pandemic in China using a small sample, which 
only represented local characteristic of the outbreak \cite{yang}. Moreover, their training dataset is the 2003 SARS epidemic statistics, which is different from COVID-19 epidemic.

Recently, an artificial neural network (ANN) has been given attention after the success of deep learning on image classification \cite{krizhevsky}. In particular, for prediction or forecasting, researchers were re-exploring the old models of ANN for time series prediction such as recurrent neural network (RNN) and LSTM. The return of ANN aims to solve the drawback of statistical methods. It performs better than statistical methods in terms of prediction accuracy \cite{zhang}. Time-series data that contain dynamic information over time are suitable to be captured by the RNN family. One special property of the RNN family is that the activation of every timestamp is stored in the internal state to construct a temporal model \cite{bayer}. However, the weakness of RNN is dealing with long-sequence data insisting the inability to handle the vanishing gradient problem during the learning process\cite{pascanu}. To solve this problem, Schmidhuber has proposed the LSTM, which contains the input, output, and forget gate to better capture the correlation of data with long-term dependencies \cite{hochreiter}. The LSTM parameter, however, needs to be optimized depending on data characteristics by choosing the number of the layers or hidden units, especially for highly complex data, which are non-linear and long \cite{langkvist}.

In this paper, we propose an LSTM framework that handle the nonlinearity and complexity of COVID-19 time-series data. The LSTM framework contains layers of LSTM cells, followed with sigmoid activation and dropout regularization. Each LSTM layer handles different resolutions of temporal space for specific tasks. Input information is forwarded through layers until the linear layer produces a time-series output. In specific, this framework is run to solve the regression problem. We can gradually add layers and hidden units to increase connections between hidden units horizontally and vertically and to improve accuracy depending on the complexity of distribution in the dataset. It captures temporal dynamics hierarchically and sequentially on complicated and long-sequence data\cite{langkvist}.

We have prepared a learning scenario that can train COVID-19 spread over time. We split training and testing data from each country for all samples. Specifically, the sequence of the selected country is split into input training and output training or label. The best LSTM architecture and hyper-parameters are searched heuristically during validation. For evaluation, we also compare the LSTM model with the precedent model of RNN.

The paper is organized as follows. Section 1 presents an introduction of the study. Section 2 elaborates our motivation of applying LSTM  to predict COVID-19 growth. Section 3 describes the methodology used in this research starting from preprocessing, learning algorithm, training, and validation strategy. Section 4 presents the experimental results. Section 5 provides discussions. Finally, Section 6 displays the conclusion of the study.

\section{Related Works and Motivation}
COVID-19 growth data include temporal information presenting the dynamic number of confirmed infected people over time. Thus, it is important to check whether the policy undertaken is effective or not during a pandemic. Studying about how to effectively treat pandemics by looking into previous and global patterns can also be performed. Moreover, in real-time, the end of pandemic can be suitably predicted given the abundance of available training data. However, by its nature characteristics, COVID-19 time-series data are complex, highly nonlinear, long interval (several days and months), and high variance making it difficult for traditional statistical methods to predict \cite{taieb}. Furthermore, the use of several hidden layers and nonlinearity are advantageous in terms of graph accuracy by capturing the coarse and fine dynamics of the growth pattern \cite{lecun}.

The use of multivariate datasets as a data sources for training the model is beneficial because pandemic growth is influenced by many factors. That is, the cause of confirmed cases can be seen from several parameters, not only stand-alone variables. In this case, for a preliminary, number of confirmed cases, death, recovered, latitude, and longitude are used as parameters. There are relationships between geographical parameters like latitude and longitude with the number of confirmed cases in the world based on previous findings \cite{yudistira}. In the future, it can be more beneficial to add new parameters such as the UV index, humidity, and population density.

The use of LSTM to overcome the drawback such as nonlinearity, long series, and heterogeneous properties, basically starts from the RNN problem. The main drawback of RNN is vanishing gradients, which can be handled by LSTM. Due to the use of the hyperbolic tangent as the activation function, the derivative of the function inside RNN cell is in the range of 0 to 1. If the gradient is very small, then there is no effect on the update.

\section{Methods}
\subsection{Data Preprocessing}

We use a MinMaxScaler to normalize data because LSTM is very sensitive to normalization, especially for capturing time-series data. First, we transform data into the same scale and thus avoid bias during training and validation. The scaling function is defined as

\begin{equation}
X_{scaled} =\frac{X-X_{min}}{X_{max}-X_{min}}
\end{equation}

where $X$ is input training dataset and $X_{scaled}$ is output of normalized training dataset.

\begin{figure*}[h!]
\includegraphics[width=1.0\textwidth]{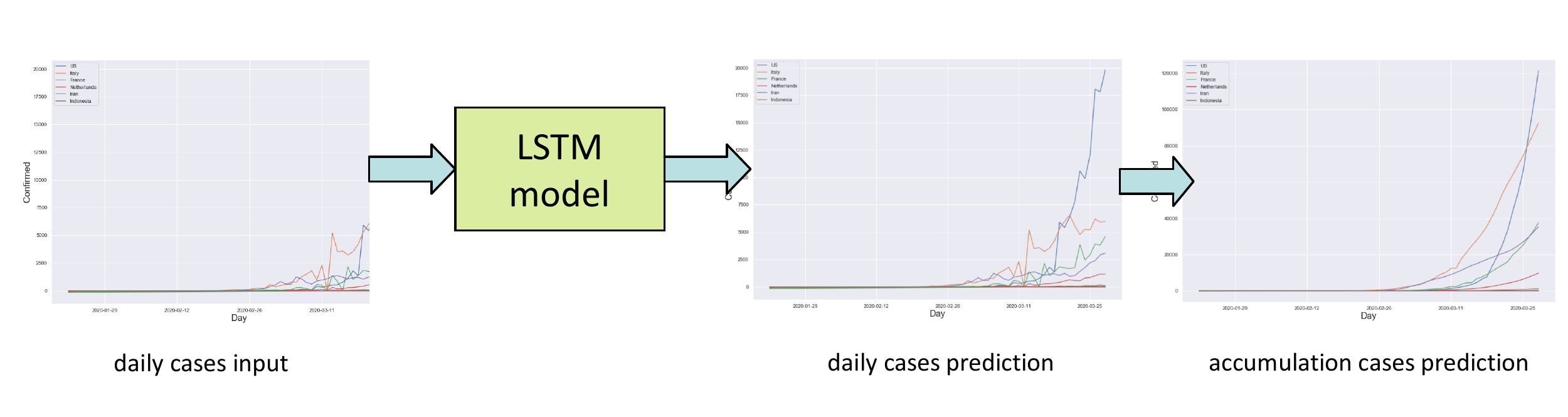}
\caption{\label{fig:framework}Framework of training and testing. The input is composed of 67 timestamps and the output is composed of 100 timestamps of daily cases. The final output is then accumulated sequentially. }
\end{figure*}

\subsection{Model}
\begin{figure}[!htb]
\includegraphics[width=.5\textwidth]{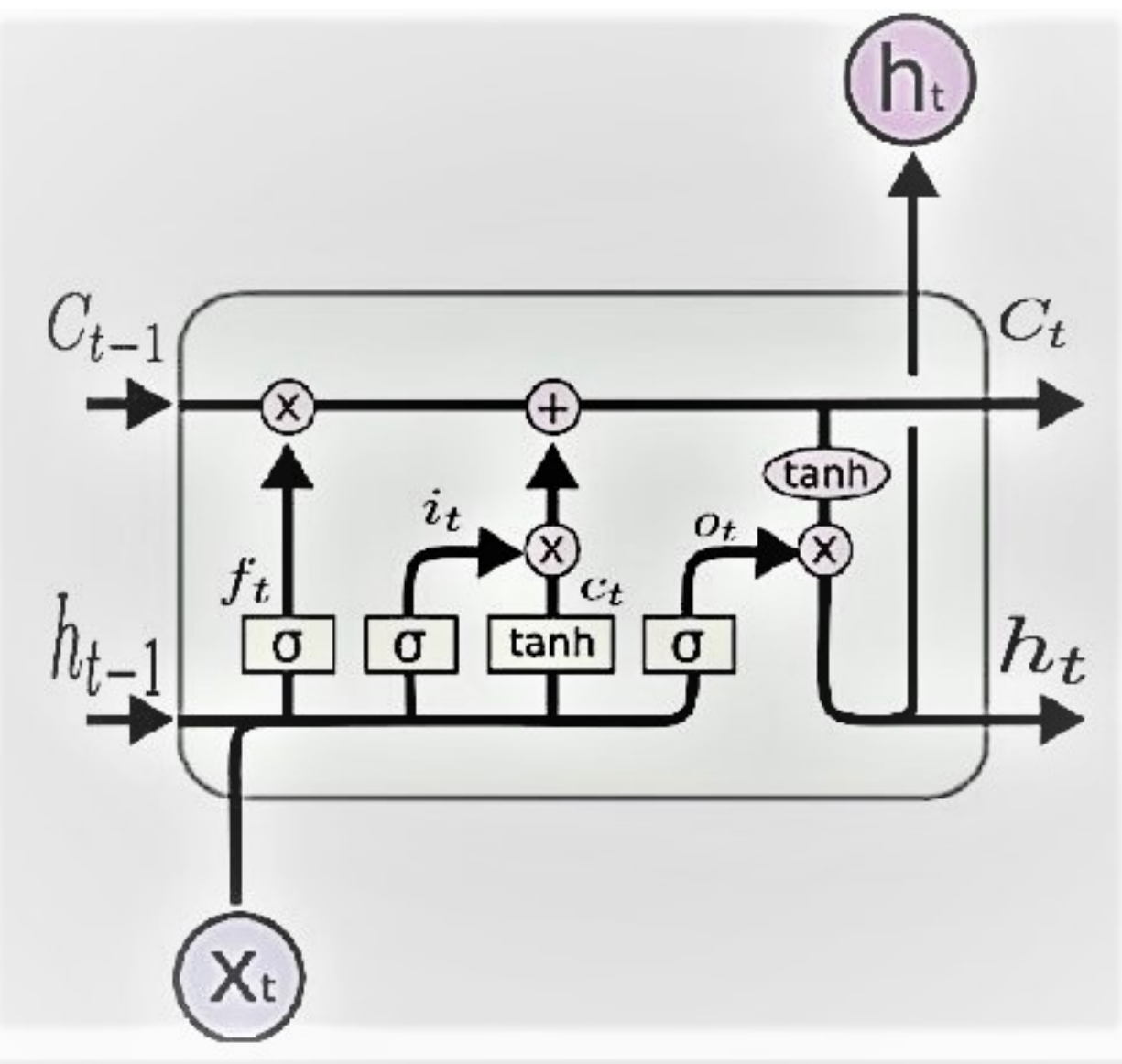}
\caption{\label{fig:lstmcell}Memory Cell.}
\end{figure}

LSTM is an extension of RNN that uses a forgetting mechanism to handle long-sequence inputs. In the LSTM cell, the memory cell is divided into memory cell $c_t$ and working cell $h_t$. Memory cells are responsible for the retention of the sequence controlled with forgetting gate $f_t$. The working memory $h_t$ is used as the output of each memory cell, and output gate $o_t$ controls the portion of $c_t$ to be remembered. The input gate $i_t$ controls the portion the former state $h_{t−1}$ and the current input $x_t$ to be remembered in the memory cell. The former state $h_{t−1}$ and current input $x_t$ are jointly fed to the non-linear activation function $tanh$ and thus not static even after a linear combination. The previously described LSTM cell shown in Figure \ref{fig:lstmcell} is elaborated as follows:

\begin{equation}
\begin{array}{l}
f_t ={ \sigma(w_f × [h_{t−1}, x_t] + b)}\\
i_t ={ \sigma((w_i × [h_{t−1}, x_t] + b_i)}\\
C_t ={ tanh(w_c \times [h_{t−1}, x_t] + b_c))}\\
c_t ={  f_t \times c_{t−1} + i_t \times c_t}\\
o_t ={ \sigma(w_o × [h_{t−1}, x_t] + b)}\\
h_t ={ o_t \times tanh(c_t)}\\
\end{array}
\end{equation}

\begin{figure}[!htb]
\includegraphics[width=.5\textwidth]{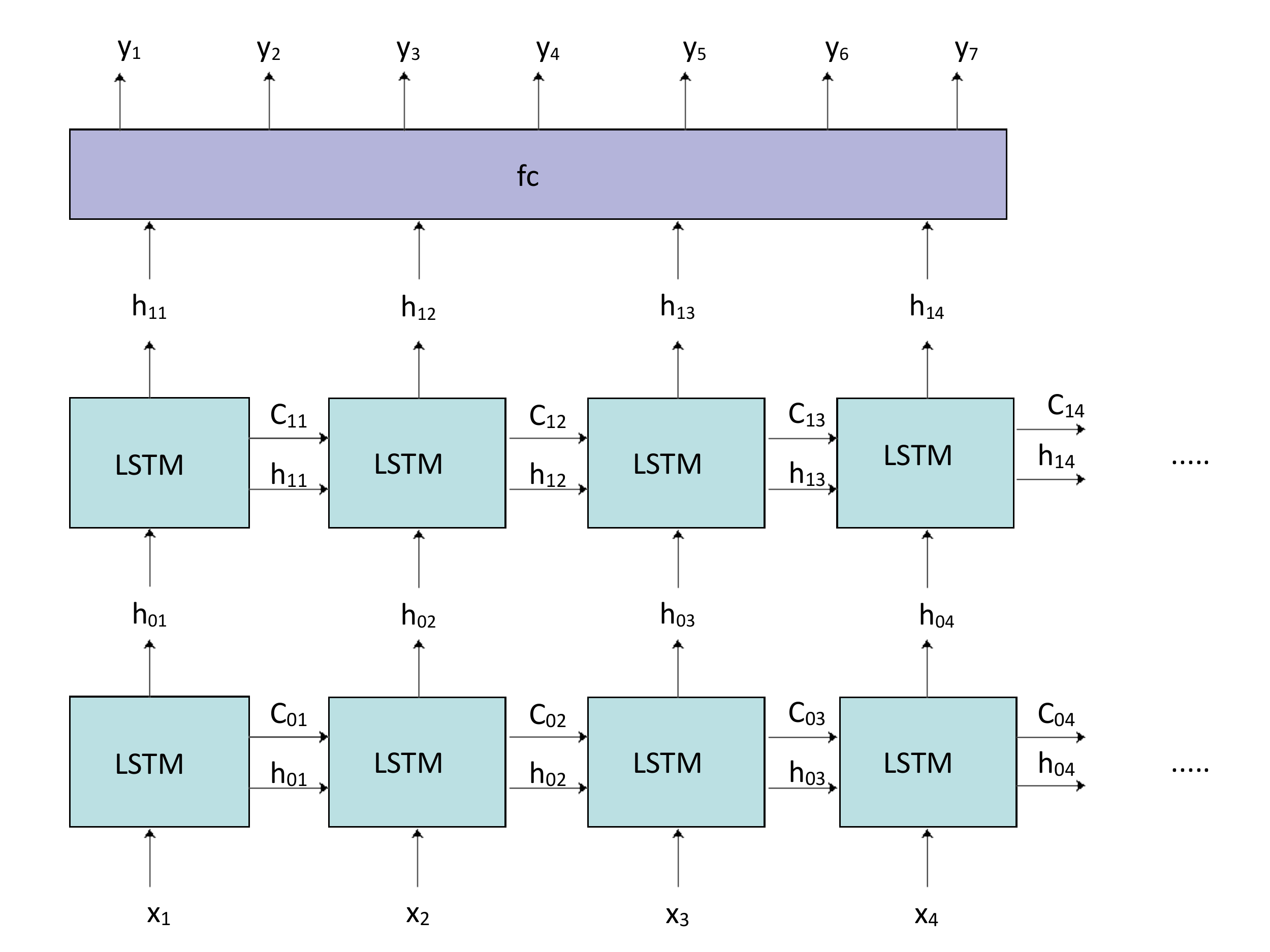}
\caption{\label{fig:lstm2layer}LSTM model architecture with two hidden layers and fully connected layer (fc).}
\end{figure}

Our architecture contains one to four hidden layers with a hidden unit of 1-30 each. An example of an LSTM architecture with two hidden layers is shown in Figure \ref{fig:lstm2layer}. The last layer is a linear layer that outputs 100-sequence prediction. The output of the linear layer is fed to the activation function of sigmoid to guarantee a range of 0 to 1. We use a dropout of 0.1 to avoid overfitting.

As shown in Figure \ref{fig:framework}, the framework of learning and evaluation consists of an input, fed into the model, and output. The input and output are both in the form of daily cases. The result of daily cases is then finally accumulated to show the growth curve over time. In the training phase, a 100-sequence input is split into the $1^{st}$ to $67^{th}$-day as input graph and $68^{th}$ to $100^{th}$-day as the label. Input is normalized before processing using the normalization. The validation and testing data are normalized using the scaling factors obtained from training data before feeding them into the trained model.

\begin{table}[h]
\centering
\scalebox{0.9}{
\begin{tabular}{| c | c | c|}
\hline
\textbf{Training data} &  \textbf{Validation Data} 
\\
\hline
'China*','Germany','Australia*','Brazil', & 'Indonesia','Sweden',\\'US','Belgium','Spain','Italy','France*',&'Saudi Arabia','Argentina'\\'Malaysia','Vietnam','Iran','UEA',&\\'Singapore','Thailand','Korea, South',&\\'Japan','Iran',&\\'Netherlands*','Russia','Chile',&\\'India','Greece','Mexico',&\\'Mongolia','Philippines','New Zealand',&\\'South Africa','Botswana','Uruguay',&\\'Paraguay','Madagascar','Peru', 'Portugal',&\\ 'Denmark*','Hungary','Kenya','Ireland','Israel',&\\'Norway','Mauritius','Rwanda','Iceland',&\\'Kazakhstan','Switzerland','Cyprus','Zimbabwe'&\\   
\hline
\end{tabular}
}
\caption{\label{traintest}Composition of training and testing data. *) has more than one province or state}
\end{table}

\subsection{Training Data}
We use 100 regions (countries/provinces/states) as the training data and four countries as validation data. The composition training and validation of selected countries are shown in Table \ref{traintest}. the parameters of the dataset are shown in Table \ref{param}. To provide the input and label, a sequence for each sample is divided into two parts, of which the first part is from January 22, 2020 to March 29, 2020 as the input training and from March 30, 2020 to May 1, 2020 as output label.
\begin{table}[h]
\centering
\scalebox{0.9}{
\begin{tabular}{| c | c | c|}
\hline
\textbf{Dataset Parameter} &  \textbf{Unit} 
\\
\hline
Confirmed cases  & number of people  \\
Death cases  & number of people \\
Recovered cases   & number of people\\
Latitude  &  degree \\
Longitude & degree\\
\hline
\end{tabular}
}
\caption{\label{param}Parameters}
\end{table}
\subsection{Prediction Accuracy Measurement}

To measure the loss function and prediction performance of the trained model, the mean squared error and root mean squared error (RMSE), respectively, are employed. Booth mean squared error and RMSE are basically the measurement of the difference between the actual cases and prediction. The RMSE is given by

\begin{equation}
RMSE ={ \sqrt{\frac{1}{N}{\sum_{i=1}^{N}{(P-A)^{2}}}}}
\end{equation}

Note that $P$ isthe  prediction sequence and $A$ is the actual or ground-truth sequence.

\section{Experimental Results}

For the hyper-parameters, we used Adam optimizer with a learning rate of 0.001 and iteration number of 10,000. These settings gave pleasant results. We have prepared several experimental setups. First, training and testing are performed once to predict a long growth curve starting from January 22, 2020  to May 1, 2020. Second, training and testing are performed five times to reduce the bias due to random initialization. In this way, we can also gather the mean and interval from several curve predictions. Meanwhile, quantitative evaluation is done using RMSE. Each country is evaluated by RMSE in five trials. We also make an evaluation on how the number of hidden states influences RMSE. Furthermore, an evaluation of the optimum number of hidden layers was conducted, in which we used a fixed number of 30 hidden states. Finally, we test to foresee when the number of daily confirmed cases is decreasing. To realize this, we predict the future growth of confirmed cases given the last 67 days of known time-series input. The best number of hidden states and layers revealed from the validation are set for testing, and a sample country of Indonesia is used as the testing input. 

\begin{figure}[!htb]
\includegraphics[width=.5\textwidth]{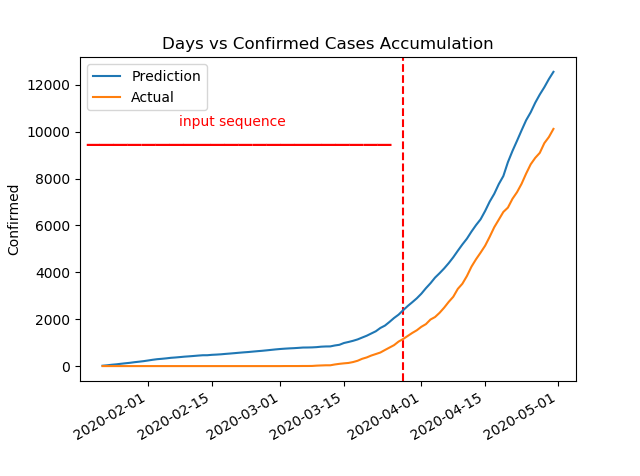}
\caption{\label{fig:indo1}Sample of prediction of Indonesia's confirmed COVID-19 cases from January 22, 2020 to May 1, 2020.}
\end{figure}

\subsection{Validation Results}

Figure \ref{fig:indo1} shows the validation results of Indonesia. The prediction curve has an exponentially similar pattern to the actual growth. The prediction is ahead of several days than the actual. The prediction on May 1, 2020 shows the number of confirmed cases, which is more than 12,000. However, in actual growth, the number of confirmed cases is still more than 10,000. This small gap is not considered significant, and it can be revealed that the daily reported cases are still on track with the reported cases worldwide. In the training data, there are various COVID-19 human test sampling that has been performed by several countries. For instance, in the US, the test sampling has already been above 1,000,000, whereas that in several other countries is still below 1,000 \cite{world}.

\begin{figure}[!htb]
\includegraphics[width=.5\textwidth]{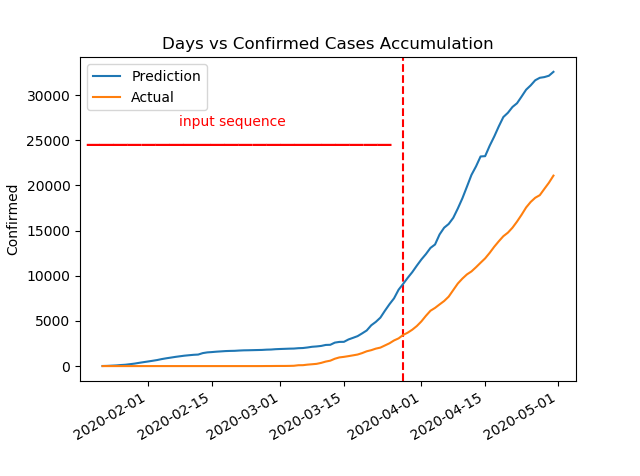}
\caption{\label{fig:swe1}Sample of prediction of Sweden's confirmed COVID-19 cases from January 22, 2020 to May 1, 2020.}
\end{figure}

Figure \ref{fig:swe1} shows the validation results of Sweden. The country is a northern subtropical country and is also well known for implementing light restrictions during the COVID-19 pandemic by only identifying the aged people as the at-risk group. The prediction shows that the number of cases grows exponentially higher than  the actual cases. However, it has  quite a similar slope with the actual  prediction. The prediction for May 1,  2020 shows that the confirmed cases reach around 30,000, greater than the actual cases, which still reaches is more than 20,000 \cite{world}.

\begin{figure}[!htb]
\includegraphics[width=.5\textwidth]{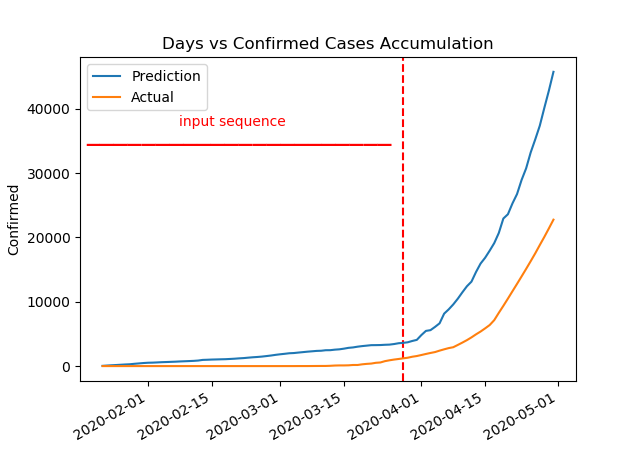}
\caption{\label{fig:arab1}Sample of prediction of Saudi Arabia's confirmed COVID-19 cases from January 22, 2020 to May 1, 2020.}
\end{figure}

Figure \ref{fig:arab1} shows the validation results of Saudi Arabia. Saudi Arabia is a tropical country similar to Indonesia, but it has higher confirmed cases. The prediction is quite similar to actual prediction in terms of the exponential curve. However, in terms of quantity, there is a significant gap where the prediction reaches more than 40,000, and the actual growth is still more than 20,000.

\begin{figure}[h!]
\includegraphics[width=.5\textwidth]{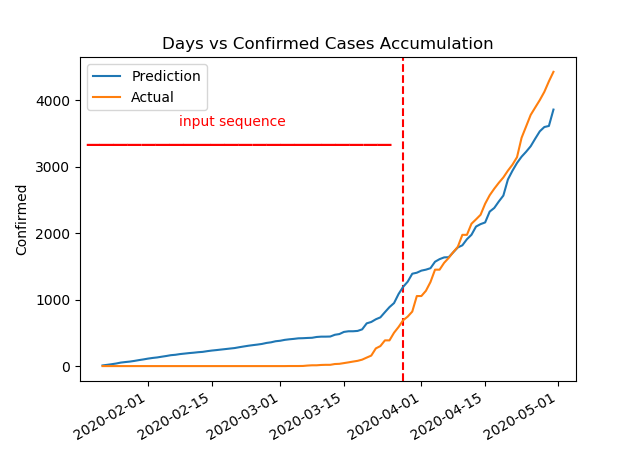}
\caption{\label{fig:meanargen}Sample of prediction of Argentina's confirmed COVID-19 cases from January 22, 2020 to May 1, 2020.}
\end{figure}

Figure \ref{fig:meanargen} shows validation results of Argentina. Argentina is a southern subtropical country. The prediction curve interchanges with the actual growth over time and grows exponentially. The prediction and actual growth reach around 4,000 on May 2, 2020.

\subsection{Interval and Mean Validation}

We also investigate the interval and mean validation of the interval and mean validation of the training and testing data for five times. This evaluation is set due to the randomness of the initial weight, making it advantageous to output several possibilities of the prediction curve. The output validation can be categorized into best, normal, and worst-case depending on the final accumulation of confirmed cases. The normal case is an average of five times of the training and validations. The best case is a graph that achieves the lowest number of accumulations of confirmed cases on June 2, 2020 and vice versa.

\begin{figure}[h!]
\includegraphics[width=.5\textwidth]{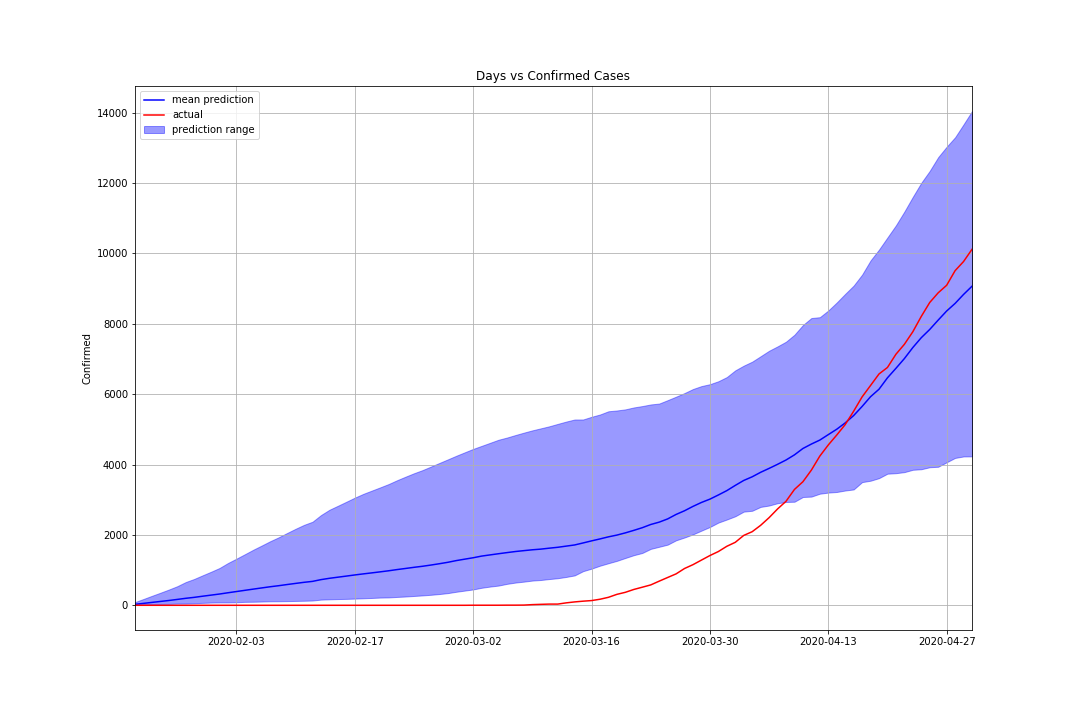}
\caption{\label{fig:indonesiamean}Sample of the mean prediction of Indonesia's confirmed COVID-19 cases from January 22, 2020 to May 1, 2020.}
\end{figure}

Figure \ref{fig:indonesiamean} shows the mean validation results of Indonesia. The actual prediction starts from the lower part of the prediction curve and gradually passes the prediction curve. The final actual growth is still within the range of prediction area. The evaluation result shows the mean RMSE is ,1,111.52, as shown in Table \ref{countries}.

\begin{figure}[h!]
\includegraphics[width=.5\textwidth]{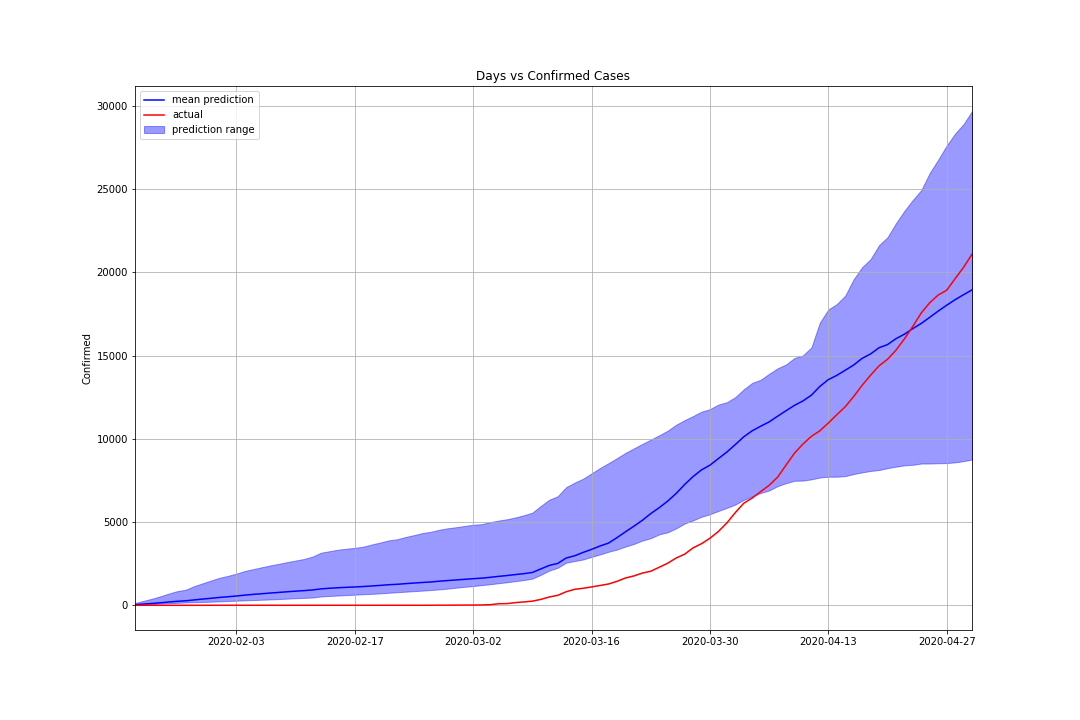}
\caption{\label{fig:swedenmean}Sample of the mean prediction of Sweden's confirmed COVID-19 cases from January 22, 2020 to May 1, 2020.}
\end{figure}

Figure \ref{fig:swedenmean} shows mean validation results of Sweden. The actual prediction starts from the lower part of prediction curve and gradually passes the prediction curve. The final actual growth is still within the range of the prediction area, with a mean RMSE of 1,756.58 (Table \ref{countries}).

\begin{figure}[h!]
\includegraphics[width=.5\textwidth]{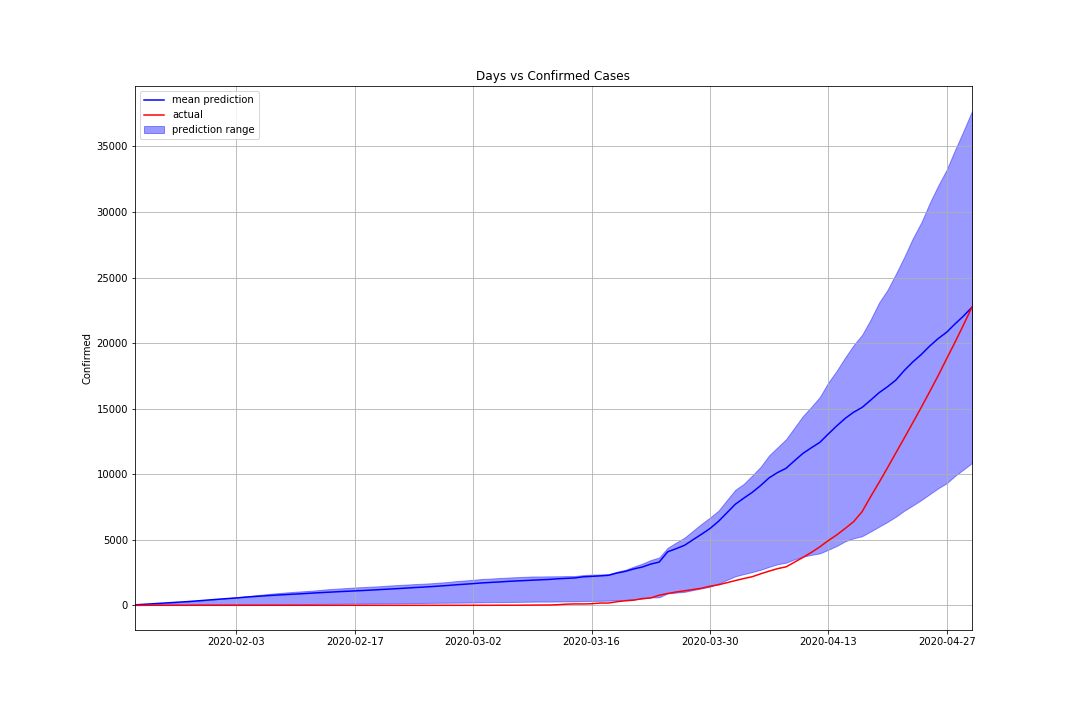}
\caption{\label{fig:saudimean}Sample of the mean prediction of Saudi Arabia's confirmed COVID-19 cases from January 22, 2020 to May 1, 2020.}
\end{figure}

Figure \ref{fig:saudimean} shows the mean validation results of Saudi Arabia. The actual curve starts from the lower part of prediction curve and finally achieves the same number of accumulated confirmed cases with the prediction. The final prediction is still within the range of the prediction areas with a mean RMSE of 2,795.88 (Table \ref{countries}).

\begin{figure}[h!]
\includegraphics[width=.5\textwidth]{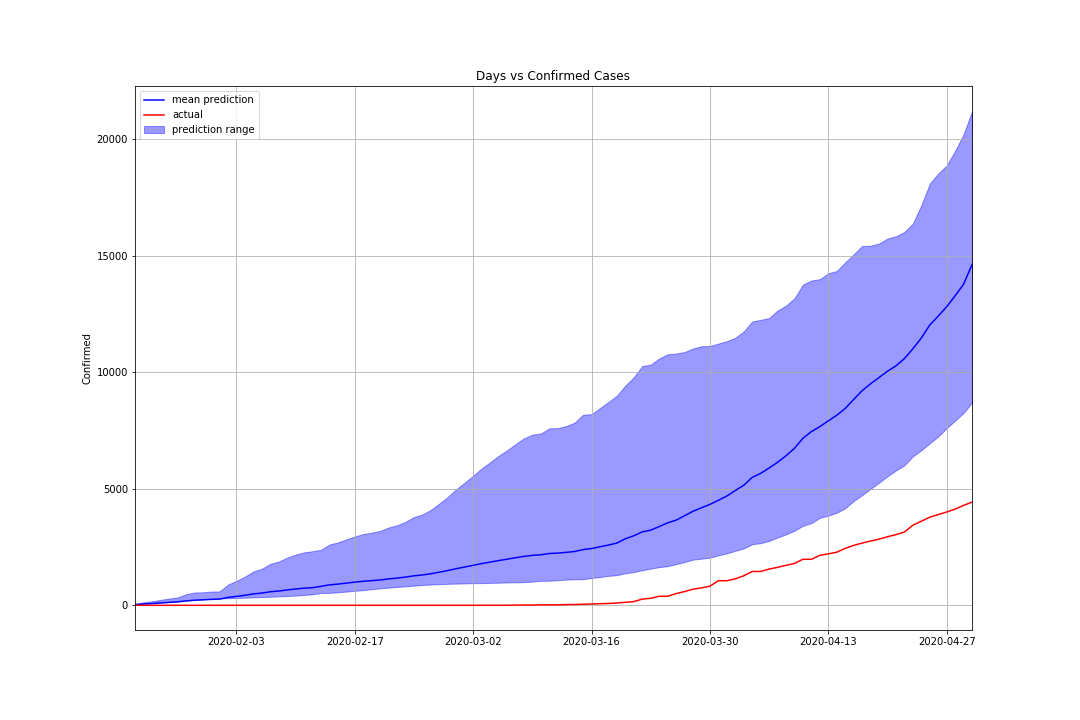}
\caption{\label{fig:argenmean}Sample of the mean prediction of Argentina's confirmed COVID-19 cases from January 22, 2020 to May 1, 2020.}
\end{figure}

Figure \ref{fig:argenmean} shows the mean prediction results of Argentina. The actual prediction starts from lower part of the prediction curve and the gap becomes wider over time. The final prediction is still outside the range of the prediction areas with a mean RMSE of 3,691.23 (Table \ref{countries}). This result regards the importance of the initial weight until achieving the best validation results. Another factor is the sample imbalance where the number of southern subtropical countries is less than that of northern subtropical and tropical countries.

\begin{figure*}[h!]
\includegraphics[width=1.\textwidth]{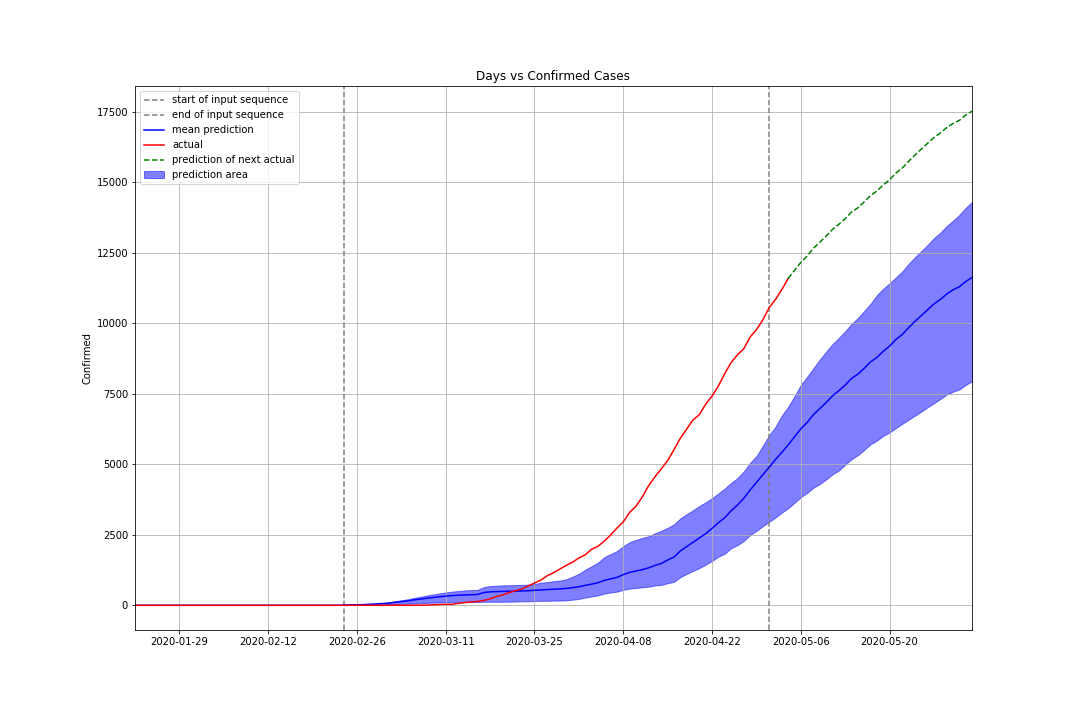}
\caption{\label{fig:predaccum}Sample of prediction of accumulation of confirmed COVID-19 cases in Indonesia from April 22, 2020 to June 2, 2020. }
\end{figure*}

\begin{figure*}[h!]
\includegraphics[width=1.\textwidth]{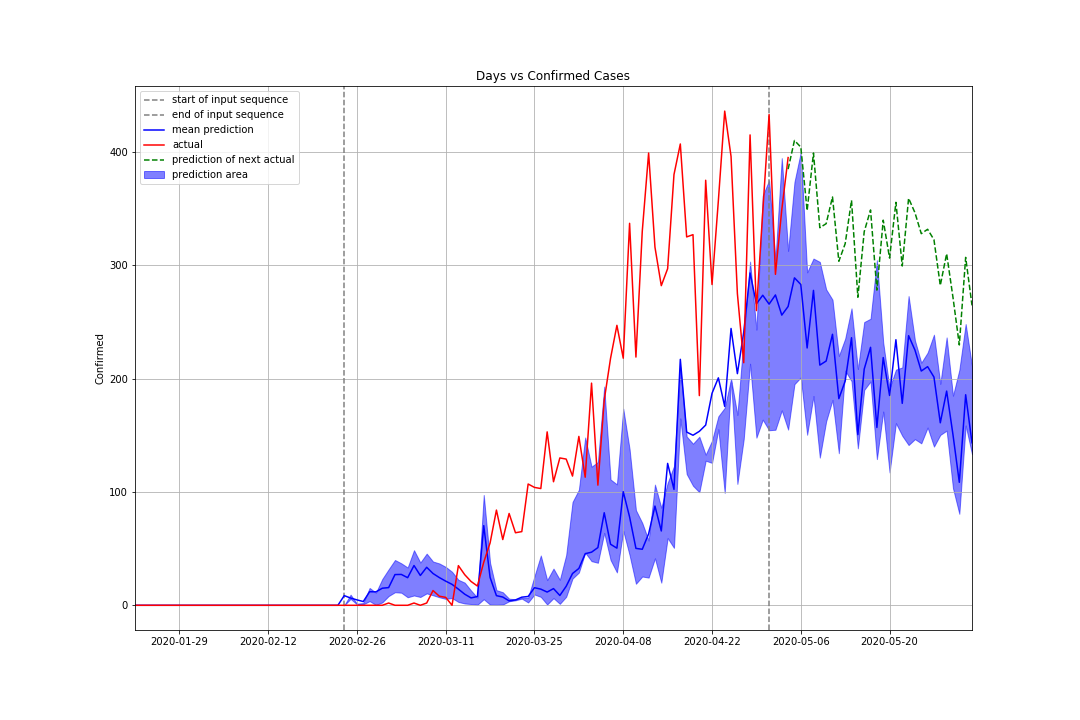}
\caption{\label{fig:preddaily}Sample of prediction of daily confirmed COVID-19 cases in Indonesia from April 22, 2020 to June 2,  2020.}
\end{figure*}

\begin{table}[h]
\centering
\scalebox{0.9}{
\begin{tabular}{| c | c | c|}
\hline
\textbf{Country} &  \textbf{RMSE} 
\\
\hline
Indonesia  & 1111.52  \\
Sweden  & 1756.58 \\
Saudi Arabia   & 2795.88\\
Argentina  & 3691.23 \\
\hline
\end{tabular}
}
\caption{\label{countries}Accuracy result of each country}
\end{table}

\begin{table}[h]
\centering
\scalebox{0.9}{
\begin{tabular}{| c | c | c|}
\hline
\textbf{Number of hidden states} &  \textbf{RMSE} 
\\
\hline
1  & 4517.87  \\
5  & 1284.94  \\
10  & 924.89 \\
30  & 889.44 \\
\hline
\end{tabular}
}
\caption{\label{states}Accuracy results given with various numbers of hidden states using four LSTM layers}
\end{table}

\begin{table}[h]
\centering
\scalebox{0.9}{
\begin{tabular}{| c | c | c|}
\hline
\textbf{Number of hidden layers} &  \textbf{RMSE} 
\\
\hline
1  & 3004.28  \\
2  & 654.22  \\
3  & 641.93 \\
4  & 568.35 \\
\hline
\end{tabular}
}
\caption{\label{layers}Accuracy results given various numbers of hidden layer using 30 hidden states}
\end{table}

\begin{table}[h]
\centering
\scalebox{0.9}{
\begin{tabular}{| c | c | c|}
\hline
\textbf{Architecture} &  \textbf{RMSE} 
\\
\hline
RNN  & 1520.61  \\
LSTM  & 1238.66 \\
\hline
\end{tabular}
}
\caption{\label{methods}Accuracy results of RNN and LSTM}
\end{table}

\subsection{Effect of Different Architectures}

We try on a different number of hidden states in each LSTM layer. The higher the number of hidden states, the higher parameter will be available inside the model. Over-parameterization will waste computational power and is thus inefficient for applications. Under-parameterization will reduce the prediction accuracy than it should be. The optimum number of the parameter is somewhat more desirable, and thus we heuristically add the number of hidden states and layers  to examine the effect on prediction accuracy. As shown in Table \ref{states}, with four hidden layers, as the number of hidden states increases, the RMSE decreased. Thus, with 30 hidden states, the LSTM model still produces significant accuracy. In Table \ref{layers}, with 30 hidden states, as the number of layers increases, the performance of the LSTM model increases.

\subsection{Comparison with RNN}

We compare our LSTM with the previous version of the time-series prediction model of RNN. As shown in Table \ref{methods}, using one-layer LSTM or RNN, LSTM outperformed RNN by 281.95. This finding confirms the ability of LSTM to recognize a long series by minimizing vanishing gradients.

\subsection{Testing to predict the future growth of COVID-19 cases until June 2, 2020}
We also arrange a  real prediction using the aforementioned LSTM model on training data from January 22, 2020 to May 1, 2020. It is then tested on the 
input sequence with a duration from February 29, 2020 to May 1, 2020. As shown in Figure \ref{fig:predaccum}, the predicted and actual curve grow with significantly different quantities, but the prediction pattern still follows the same exponential curve. We confirm this assumption by looking into the daily confirmed cases (Figure \ref{fig:preddaily}). The blue one has the same growth pattern with the daily cases, presented by the red lines (actual), but with different quantities. This finding shows the ability of LSTM to capture growth pattern more than the quantity. We suggest that more data should be included in the training phase for more precise results. To predict the continuation of the actual graph (red),  the portion of the blue graph (mean prediction) is cut starting from the end of the actual graph (May 2, 2020).  Its cut series is then uniformly augmented, such that it is at the same level as the actual graph. The final continuation prediction shows the decreasing trend in May with the range of cases between 400 to 300 and fbelow 300 cases after May 20, 
2020. This cut and augmentation method can be performed daily to update the prediction. 

\section{Discussions}

LSTM is a model that captures the correlation of time series dynamics. This research verifies the ability of LSTM to predict the COVID-19 growth curve given enough training data. The results will be convincingly better if we add more variety of data with large quantities (big data). Our approach is absolutely better than the traditional statistical approach or qualitative modeling because the model is trained to represent global data optimally. The samples used to train LSTM is divided by 67 days after January 22, 2020 as the input and 33 days before May 1, 2020 as the output  The total sequence is 100 days. The input of the training and testing and output are 67-sequence and 100 days, respectively. Therefore, long-term COVID-19 growth prediction is a difficult problem due to long-range time-series prediction (many-to-many).

Regarding the parameters employed in this research, the latitude and longitude represent the confirmed cases well and show that northern subtropical countries tend to have a steeper growth slope than the tropical and southern ones. This conclusion is drawn quantitatively from the RMSE results in the validation phase.

\section{Conclusion}

We have developed an LSTM based prediction model to foresee the COVID-19 pandemic growth over countries. The COVID-19 data are time-series data of which the accumulated number of confirmed COVID-19 cases is monotonically increasing over time until it arrives at a certain converged peak curve. Given large training data, LSTM capture the pattern of the dynamic growth of graphs with a minimum RMSE compared to RNN. The results suggest that LSTM is a promising tool to predict the COVID-19 pandemics by learning from big data and can potentially predict future outbreaks. Future work should increase the training data by either adding new data or a data augmentation strategy.


%



\section*{Acknowledgment}
The authors of this paper would like to express their
thanks and gratitude to Sutiman Bambang Sumitro from the Department of Biology, Faculty of Mathematics and Natural Science, Brawijaya University for his support and guidance in this COVID-19 research.

\ifCLASSOPTIONcaptionsoff
  \newpage
\fi

\end{document}